\documentclass{article}
\usepackage{ijcai25}

\usepackage{times}
\usepackage{soul}
\usepackage{url}
\usepackage[hidelinks]{hyperref}
\usepackage[utf8]{inputenc}
\usepackage[small]{caption}
\usepackage{graphicx}
\usepackage{amsmath}
\usepackage{amsthm}
\usepackage{booktabs}
\usepackage{algorithm}
\usepackage{algorithmic}
\usepackage[switch]{lineno}

\newcommand{\Not}{\textit{not}\,}
\newcommand{\head}{\textit{head}}
\newcommand{\body}{\textit{body}}
\newcommand{\HU}{\textit{HU}}

\newcommand{\LF}{\textit{LF}}
\usepackage{multirow}
\usepackage{listings}
\usepackage{comment}
\usepackage{parskip}
\usepackage[dvipsnames]{xcolor}
\usepackage{rotating}
\usepackage{booktabs}
\usepackage{makecell}
\usepackage{multirow}
\usepackage{makecell}
\usepackage{subcaption}

\usepackage{tcolorbox}
\tcbuselibrary{listings, breakable}

\newtheorem{property}{Property}
\DeclareCaptionStyle{ruled}{labelfont=normalfont,labelsep=colon,strut=off}

\DeclareCaptionStyle{ruled}{labelfont=normalfont,labelsep=colon,strut=off}

\floatstyle{ruled}
\newfloat{listing}{tb}{lst}{}
\floatname{listing}{Listing}

\newtheorem{theorem}{Theorem}
\newtheorem{definition}{Definition}

\title{Improving Efficiency of Answer Set Planning with Rough Solutions\\ from Large Language Models for Robotic Task Planning}

\author{
Xinrui Lin$^1$ \and
Yangfan Wu$^1$ \and
Huanyu Yang$^1$ \and
Yuting Huang$^1$\and
Yu Zhang$^{1,2}$\and
Jianmin Ji$^{1,2}$\thanks{Corresponding author} \And
Yanyong Zhang$^{1,2}$ \\
\affiliations
$^1$University of Science and Technology of China, China \\
$^2$Institute of Artificial Intelligence, Hefei Comprehensive National Science Center, China \\
\emails
\{xinruilin, uuyf, yanghuanyu, yutinghuang\}@mail.ustc.edu.cn, \\
\{yuzhang, jianmin, yanyongz\}@ustc.edu.cn
}

\begin{document}

\maketitle
\begin{abstract}
Answer Set Programming (ASP) planning can be used to refine the rough solutions generated by Large Language Models (LLMs) to handle specific restrictions of actions, i.e., reconstruct the rough solutions to be executable, for robotic task planning.
However, it is still challenging to efficiently solve ASP programs that have multiple variables with large domains, which prevents the above application of ASP planning from real-world task planning problems. In this paper, we consider how to reduce the domains of variables without losing possible solutions for ASP planning, while given these rough solutions from LLMs.
Based on the above reduction, we introduce CLMASP, an approach that couples LLMs with ASP for robotic task planning.
We evaluate CLMASP on the VirtualHome platform for common indoor tasks, demonstrating a significant improvement in the executable rate from under 10\% to nearly 90\% and reducing average ASP planning time from over 2 hours to under 5 seconds.
Code is available at \url{https://github.com/CLMASP/CLMASP}.
\end{abstract}

\section{Introduction}

Answer Set Programming (ASP)~\cite{lifschitz2019answer} has been applied to robotic task planning~\cite{chen2013handling,tran2023answer}. However, answer set planning still faces challenges in real-world applications due to the extensive domain knowledge required and the limited computational efficiency in large-scale problems.

Conversely, Large Language Models (LLMs), such as GPT-4~\cite{openai2023gpt4blog} and Llama 3~\cite{meta_llama_3}, possessing broad foundational knowledge, are well-suited for task planning in open-world scenarios~\cite{kambhampati2024llmscantplanhelp}. It is hard for LLMs to directly generate executable plans that follow specific constraints of the robot~\cite{wu2023embodied}. For instance, an LLM-generated plan might skip essential actions like ``\texttt{plugin washing\_machine}'' before ``\texttt{switch\_on washing\_machine}''.
However, these LLM-generated rough solutions serve as valuable guidance.

In this paper, we consider how to utilize these rough solutions to speed up answer set planning. Rough solutions from LLMs can improve the computational efficiency of answer set planning in two different ways.
Firstly, inspired by the TLPLAN system~\cite{bacchus2000using} that utilizes domain-specific control knowledge to speed up the planning process, we encode the ASP program to refine the rough plan to accomplish the task. In this sense, we define the planning problem with a rough solution and provide an ASP encoding to generate executable plans for the problem.
Secondly, we consider how to reduce the domains of variables to speed up the grounding process in ASP. However, the ground ASP program obtained by a reduction may generate different results from the original program.  To address this,  we introduce the notions of \emph{admissible reduction} and \emph{safe reduction} that preserve possible solutions for the original program, while \emph{admissible reductions} are more helpful for answer set planning.  It is hard to identify \emph{admissible} or \emph{safe reductions}. Then we propose a sufficient condition with the notions of loops and loop formulas~\cite{lin2004assat} for identifying \emph{admissible reductions} and apply the condition in our encoding for answer set planning.

 \begin{figure}[!tbp]
     \centering
     \includegraphics[width=1.05\linewidth]{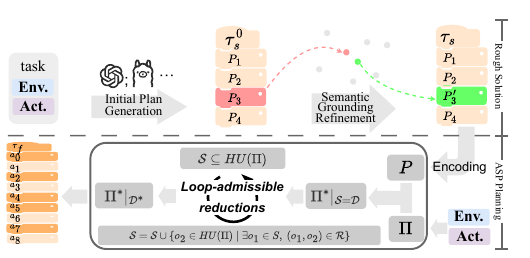}
     \caption{CLMASP workflow: An LLM generates an initial plan which undergoes Semantic Grounding Refinement to produce a rough solution, which is then enhanced by ASP to generate a detailed, executable plan.}
     \label{fig:workflow}
\end{figure}

For a typical task planning problem with hundreds of objects and requiring dozens of actions to accomplish the task, the above methods can reduce the computing time from more than 2 hours to less than 5 seconds. Building on the two speedup  methods, we introduce \textbf{CLMASP} (see Figure~\ref{fig:workflow}), an approach that couples LLMs with ASP for robotic task planning.  Our contributions are as follows:
\begin{itemize}
    \item We consider leveraging rough solutions from LLMs to accelerate answer set planning. To enhance ASP planning efficiency, we propose two methods: (1) defining the planning problem using a rough solution refined by an ASP program, and (2) introducing \emph{admissible} and \emph{safe reductions}, along with a sufficient condition based on loops and loop formulas to reduce variable domains.
    \item We provide a framework, CLMASP, that leverages the accelerated ASP planning methods to improve both computational efficiency and the executability of plans, as demonstrated through experiments on the VirtualHome platform.
\end{itemize}

\section{Related Work}

\subsection{Improve Efficiency of ASP}
Reducing domains of variables to reduce the size of ground programs has not been considered much in ASP.  Lazy grounding~\cite{dal2009gasp} requires a new ASP solver implementation, rather than using existing solvers such as clingo~\cite{gebser2019multi}. Although methods such as rewriting~\cite{pearce2004simplifying,ji2015simplifying,ji2016eliminating}, splitting~\cite{split,ji2015splitting,fandinno2023splitting}, and forgetting~\cite{ji2015forgetting,saribatur2024unified} have been extensively explored, they often require some sense of equivalence during the simplification of logic programs.
Abstraction~\cite{saribatur2021omission,saribatur2021abstraction} approximates answer sets without losing any, by simplifying logic programs so that every answer set of the original program corresponds to one in the abstract program. However, it does not explicitly consider the grounding process. In contrast, our \emph{admissible} and \emph{safe reductions} operate at the grounding level by pruning constants to reduce ground program size while preserving relevant answer sets. To our knowledge, such reductions have not been explored in ASP.

\subsection{LLMs in Robotic Task Planning}

Recent research has explored the application of LLMs in robotic task planning using prompt-based methods that integrate domain knowledge through natural language~\cite{huang2022language,song2023llm} or pseudo-Python code~\cite{singh2023progprompt,liang2023code}. However, LLMs often struggle to retain crucial information in long-horizon tasks. To address these limitations, the integration of external tools like knowledge base feedback~\cite{gou2023critic}, local checkers~\cite{chen2024autotamp}, and physical environment interactions~\cite{bhat2024grounding} has been proposed, alongside fine-tuning LLMs for specific tasks to minimize inaccuracies~\cite{xiang2024language,mu2024embodiedgpt}. Nevertheless, robotic task plans still struggle to fully conform to action models and environmental constraints, complicating the successful completion of tasks.

In contrast, a well-designed symbolic planner can yield comprehensive and interpretable results within a specific domain~\cite{pan2023logic}. Current research is exploring the integration of symbolic planners with LLMs~\cite{liu2023llm+,yang2023coupling}. Compared to our proposed CLMASP, existing approaches primarily use LLMs for translating informal to formal representations, neglecting their planning and commonsense reasoning capabilities, and fail to adapt symbolic planners for effective collaboration with large models.

\section{Preliminaries}
We first recall the definition of Answer Set Programming (ASP) with variables~\cite{van2008handbook}, focusing on key concepts used in our method. We then review loops and loop formulas~\cite{lin2004assat}, and finally discuss the ASP encoding for robot task planning~\cite{tran2023answer}.

\subsection{Answer Set Programming with Variables}
\label{subsec:Answer_Set_Programming_with_Variables}
Consider a first-order vocabulary \(\Theta = (\mathcal{P}, \mathcal{D})\), where \(\mathcal{P}\) and \(\mathcal{D}\) are finite, non-empty sets of predicates and constants, respectively. A \emph{term} is a constant from \(\mathcal{D}\) or a variable from a set of variables \(\mathcal{V}\). An \emph{atom} has the form \(p(t_1, \ldots, t_n)\) where \(p \in \mathcal{P}\) and each \(t_i\) is a term. A {\em ground atom} contains no variables. An ASP program \(\Pi\) with variables consists of rules of the form:
\[
A_0 \gets A_1, \ldots, A_m, \Not A_{m+1}, \ldots, \Not A_n.
\]
where \(A_i\) are {\em atoms} ($0\leq i\leq n$) and \(n \geq 1\). Each rule~\(r\) is also written as \(\head(r) \gets \body(r)\), where \(\body(r)=A_1\land \cdots\land A_m\land \neg A_{m+1}\land\cdots\land \neg A_n\), \(\body^+(r)=\{A_1, \ldots, A_m\}\) and \(\body^-(r)=\{A_{m+1}, \ldots, A_n\}\). \(\Pi\) is {\em ground} if all its rules are {\em ground}; \(\Pi\) is {\em safe} if each variable in rule \(r\) appears in \(\body^+(r)\).
Safety is achieved using {\em domain predicates} - unary predicates given as facts, universally true for all constants used to ground the variables. The \emph{Herbrand universe} of \(\Pi\), denoted by \(\HU(\Pi)\), consists of the set of all constants, that is, \(\HU(\Pi) = \mathcal{D}\).

A \emph{complete variable assignment} \(\sigma\) maps each variable in \(\mathcal{V}\) to a constant in \(\mathcal{D}\). For a rule \(r\), the ground rule \(r\sigma\) replaces every variable in \(r\) with its value in \(\sigma\). The set of all ground instances of \(r\) over \(\mathcal{D}\), denoted \(r|_{\mathcal{D}}\), includes all possible ground rules with constants from \(\mathcal{D}\):
\[
\begin{aligned}
r|_{\mathcal{D}} &= \{\, r\sigma \mid {} \\
&\quad \sigma \text{ is a complete variable assignment for } r \text{ over } \mathcal{D}\}.
\end{aligned}
\]

For an ASP program ~\(\Pi\), we can construct its ground program \(\Pi|_{\HU(\Pi)} = \bigcup_{r\in \Pi} r|_{\HU(\Pi)}\), referred to as the {\em grounding result}.

Given an {\em interpretation}~\(I\), i.e., \(I\) is a set of ground atoms, \(I\) {\em satisfies} a ground rule \(r\),
if \(\head(r)\in I\) whenever \(\body^+(r)\subseteq I\) and \(\body^-(r)\cap I =\emptyset\).
An interpretation \(I\) is a {\em model} of \(\Pi\), if it satisfies every ground rule in \(\Pi|_{\HU(\Pi)}\).
\(I\) is a {\em minimal} model if there is no other model \(J\) of \(\Pi\) such that \(J\subset I\).

Now, we define the answer sets of an ASP program with variables.
Given an ASP program \(\Pi\) and an interpretation~\(I\),
the Gelfond-Lifschitz reduction~\cite{gelfond1988stable} of \(\Pi\) on \(I\), written \(\Pi^I\), is obtained from \(\Pi|_{\HU(\Pi)}\) by deleting:
\begin{enumerate}
\item each rule that has \(\Not A\) in its body with \(A\in I\), and
\item all \(\Not A\) in the bodies of the remaining rules.
\end{enumerate}
For any interpretation \(I\), \(\Pi^I\) is a ground positive ASP program, then \(\Pi^I\) has only one minimal model. Thus, an interpretation \(I\) is an {\em answer set} of \(\Pi\) iff \(I\) is the minimal model of \(\Pi^I\).
An ASP program may have zero, one or multiple answer sets.
We focus here on normal rules; richer constructs follow in Sec.~\ref{subsec:EncodeOfASP}.

\subsection{Loops and Loop Formulas}
With the notions of loops and loop formulas, we can convert an ASP program $\Pi$ into a propositional formula such that an interpretation  \(I\) is an answer set of $\Pi$ if{f} it is a model of the propositional formula.
Notice that, we define the notions of loops and loop formulas on the ground ASP program $\Pi|_{\HU(\Pi)}$, as we define the notion of answer sets on $\Pi|_{\HU(\Pi)}$, which slightly differs from the first-order loop formulas in~\cite{chen2006first}.

Given an ASP program $\Pi$, the {\em positive dependency graph} of $\Pi$, written $G_\Pi$, is a directed graph whose vertices are atoms in $\Pi|_{\HU(\Pi)}$ and there is an arc from $A$ to $B$ if there is a rule $r\in\Pi|_{\HU(\Pi)}$ such that $A\in \head(r)$ and $B\in \body^+(r)$.
A set $L$ of ground atoms is a {\em loop} of $\Pi$ if the $L$-induced subgraph of $G_\Pi$ is strongly connected.
Every singleton in $\Pi|_{\HU(\Pi)}$ is also a loop of $\Pi$.

Given a loop $L$ of an ASP program $\Pi$, a ground rule
$r\in \Pi|_{\HU{(\Pi)}}$
is an {\em external support} of $L$ if $\head(r)\in L$ and $L\cap \body^+(r) = \emptyset$. We denote $R^-(L, \Pi)$ to be the set of all external support rules of $L$ in $\Pi|_{\HU{(\Pi)}}$. The {\em loop formula} of $L$ under $\Pi$, written $\LF(L, \Pi)$, is given by:
\[
\bigwedge_{A\in L} A \supset \bigvee_{r\in R^-(L, \Pi)}  \body(r).
\]

\begin{theorem}
    ~\cite{lin2004assat}\label{them:1} Given an ASP ground program $\Pi$ and an interpretation $I$. If $I$ is a model of $\Pi$, then $I$ is an answer set of $\Pi$ if and only if $I$ satisfies $\LF(L, \Pi)$ for all loops $L$ of $\Pi$.
\end{theorem}

\subsection{Encoding of Answer Set Planning}\label{subsec:EncodeOfASP}
A robot task planning problem can be encoded as an ASP program $\Pi$~\cite{tran2023answer}, where each answer set $I$ corresponds to a valid plan. $\Pi$ comprises the following components:
\begin{itemize}
    \item A set of fluents $F$ representing states, \texttt{h(f,t)} (fluent \texttt{f} is true at time step \texttt{t});
    \item A set of actions $A$ represented as \texttt{occurs(a,t)} (action \texttt{a} occurs at time step \texttt{t});
    \item Initial state $s_0$ using fluents at time 0;
    \item Action preconditions as constraints on \texttt{occurs(a,t)};
    \item Effect axioms describing how actions modify fluents;
    \item Frame axioms for unchanged fluents;
    \item Goal conditions at the final time step.
\end{itemize}

A \emph{trajectory} $\tau$ is a sequence $\langle s_0, a_1, s_1, \ldots, a_n, s_n \rangle$, where each $s_i$ is the state at time $i$ $(0 \leq i \leq n)$, and $a_j$ $(1 \leq j \leq n)$ is the action at time $j$.\footnote{In this study, only one action is performed at each time step.}
An answer set $I$ of $\Pi$ contains exactly such a trajectory, where states and actions are derived from \texttt{h(f,t)} and \texttt{occurs(a,t)} present in $I$ respectively. A trajectory $\tau$ {\em satisfies} an encoding $\Pi$ if there exists an answer set $I$ of $\Pi$ such that $\tau$ can be constructed from $I$.

We now describe a practical ASP encoding $\Pi$ for robot task planning, using \emph{clingo}'s syntax~\cite{potassco-docs}.
While our preceding discussion focused on normal rules, this encoding also employs constructs like choice rules.
To find the shortest plan, we use \emph{clingo}'s incremental mode.
Listing~\ref{lst:wash} shows an ASP encoding of the \texttt{wash} and \texttt{find} actions, structured per Algorithm~\ref{alg:aspplanning}. Traditional encoding includes $\Pi_{\mathit{domain}}$ (Lines 3-10) defining objects, initial states, and actions; $\Pi_{\mathit{step}}$ (Lines 11-22) specifying action preconditions, effects, and state transitions; and $\Pi_{\mathit{check}}$ (Lines 26-27) detailing goal constraints.
When executed, the ASP solver returns an answer set ``\texttt{occurs(find(2),1). occurs(wash(2),2).}" indicating the robot first \texttt{finds} object 2, then \texttt{washes} it.

\begin{algorithm}[tb]
    \caption{ASP-based Task Planning Encoding}
\label{alg:aspplanning}
    \textbf{Input}: \\
    \hspace*{1em}$\Pi_{\mathit{domain}}$: domain encoding (objects, initial states, maximum allowed timesteps ($T_{\mathit{max}}$), etc.).\\
    \hspace*{1em}$\Pi_{\mathit{step}}$: action encoding (preconditions, effects, etc.)\\
    \hspace*{1em}$\Pi_{\mathit{check}}$: goal-checking constraints. \\
    \textbf{Output}: Plan $\tau=\langle s_0, a_1, s_1, \ldots, s_n \rangle$.\\
    \vspace{-1em}
    \begin{algorithmic}[1]
        \STATE Set $t \leftarrow 0$  \COMMENT{Initialize time step $t$ to 0}
        \STATE Load $\Pi_{\mathit{domain}}$, $\Pi_{\mathit{step}}$, $\Pi_{\mathit{check}}$
        \WHILE{no solution found and $t < T_{\mathit{max}}$}
            \STATE $t \leftarrow t + 1$  \COMMENT{Increment time step $t$}
            \STATE Ground and solve $\Pi_{\mathit{base}} \cup \Pi_{\mathit{step}}(1..t) \cup \Pi_{\mathit{check}}(t)$
            \IF{an answer set $I$ is found}
                \STATE Extract plan $\tau$ from $\mathtt{occurs(a,t)}$ and $\mathtt{h(f,t)}$ in $I$
                \STATE \textbf{return} $\tau$
            \ENDIF
        \ENDWHILE
        \STATE \textbf{return} ``No solution found within $T_{\mathit{max}}$ steps.''
    \end{algorithmic}
\end{algorithm}

\begin{listing}[!tb]
\caption{ASP Encoding for {\tt wash} and {\tt find}. Our proposed skeleton plans (Section~\ref{subsec:asp_with_skeleton_plans}, lines 23-25, 28) replace the traditional method (Section~\ref{subsec:EncodeOfASP}, commented out in lines 22, 27), both of which are compared in our experiments.}

\label{lst:wash}
\begin{lstlisting}
#include <incmode>.
#const tmax = 30.
#program base.
num(0..tmax). time(0).
is(1,character). is(2,clothesshirt).
robot(1). object(2).
action_of(wash(O)) :- object(O).
action_of(find(O)) :- object(O).
h(hand_empty(1),0). h(same_loc(1,2),0).
task(wash(O)) :- is(O,clothesshirt), object(O).
#program step(t).
time(T) :- num(T), T==t.
1{occurs(A,t):action_of(A)}1.
h(clean(O),t) :- occurs(wash(O),t), object(O).
:- occurs(wash(O),t), not h(hand_empty(C),t-1), robot(C), object(O).
:- occurs(wash(O),t), not h(found(O),t-1), object(O).
h(found(O),t) :- occurs(find(O),t), object(O).
:- occurs(find(O),t), not h(same_loc(C,O),t-1), robot(C), object(O).
h(hand_empty(C),t) :- h(hand_empty(C),t-1), not -h(hand_empty(C),t),robot(C).
h(same_loc(C,O),t) :- h(same_loc(C,O),t-1), not -h(same_loc(C,O),t), robot(C), object(O).
h(clean(O),t) :- h(clean(O),t-1), not -h(clean(O),t), object(O).
% goal(wash(O),t):-h(clean(O),t),object(O).
goal(subtask(p,find(2),1),t) :- occurs(find(2),t).
goal(subtask(p,wash(2),2),t) :- occurs(wash(2),t).
goal(p,t) :- goal(subtask(p,find(2),1),T1), time(T1), goal(subtask(p,wash(2),2),T2), time(T2), T1<=T2, T2<=t.
#program check(t).
% :-query(t),not goal(Task,t),task(Task).
:-query(t),not goal(p,t).
#show occurs/2.
\end{lstlisting}
\end{listing}

While this encoding can scale to more complex domains by adding new rules, handling numerous constants and actions in indoor task planning can lead to inefficient planning. Section~\ref{sec:ImproveEfficiency} explores methods to enhance computational efficiency.

\section{Improve Efficiency of Answer Set Planning with Rough Solutions} \label{sec:ImproveEfficiency}
Despite advances in ASP solver efficiency, real-world robot task planning remains challenging due to the computational complexity, becoming NP-complete for bounded plan lengths~\cite{tran2023answer}. LLMs, with their extensive knowledge, can facilitate the planning process by generating rough solutions. This section shows how to utilize these rough solutions to speed up answer set planning by: (1) incorporating \emph{skeleton plans} into existing ASP encoding—an idea inspired by TLPLAN~\cite{bacchus2000using}, which uses domain-specific control knowledge to guide the planner, and (2) reducing variable domains during grounding without losing any valid plans.

\subsection{Answer Set Planning with Skeleton Plans}
\label{subsec:asp_with_skeleton_plans}
To enhance the efficiency of answer set planning, we introduce \emph{skeleton plans} (e.g., generated by LLMs) that provide guided instructions for tasks.

A \emph{planning problem with a skeleton plan} is a pair \((\Pi, P)\), where \(\Pi\) is an ASP encoding of the planning domain, and \(P\) is a skeleton plan aligned with \(\Pi\)'s signature. The signature comprises disjoint sets of action names (from ground terms in \texttt{occurs(a,t)}), fluent names (from ground atoms \texttt{h(f,t)}), and subtask names, with no circular subtask references. A \emph{skeleton plan} \(P\) is recursively defined as:
\begin{itemize}
    \item An action name \(a\);
    \item A fluent-specification \(\varphi\), formed from fluent names using propositional connectives;
    \item A subtask name \(p\);
    \item A sequence \(P_1; \ldots; P_m\), where each \(P_i\) is a skeleton plan.
\end{itemize}

A trajectory \(\tau = \langle s_0, a_1, s_1, \ldots, a_n, s_n \rangle\) satisfies \(P\) if:
\begin{itemize}
    \item \(P = a\) and \(a = a_n\);
    \item \(P = \varphi\) and \(s_n\) satisfies \(\varphi\);
    \item \(P = p\) and \(\tau\) satisfies the skeleton plan for subtask \(p\);
    \item \(P = P_1; \ldots; P_m\), and there exist \(0 \leq n^1 \leq \cdots \leq n^{m-1} \leq n\) such that each sub-trajectory \(\langle s_{n^{i-1}}, a_{n^{i-1}+1}, \ldots, s_{n^i} \rangle\) satisfies \(P_i\) (with \(n^0 = 0\), \(n^m = n\)).
\end{itemize}
A trajectory \(\tau\) is a \emph{solution} to \((\Pi, P)\) if it satisfies both \(\Pi\) and \(P\). This implies that a plan for $\Pi$ can be constructed by extending the skeleton in $P$ into an executable plan, rather than exploring all possible action sequences, enhancing efficiency.

For example, a skeleton plan for task “wash object 2” can be:
\[
P = \texttt{find(2);}~\texttt{wash(2)}.
\]
This plan encodes a high-level instruction that the robot should first \texttt{finds} object 2, then \texttt{washes} it.   Although execution details are abstracted away, it still constrains the search space and is embedded into ASP via auxiliary rules.

Skeleton plans are incorporated by adding ASP rules for subtask definitions and plan satisfaction constraints, as shown in Listing~\ref{lst:wash} (Lines 23--25, 28). This enables efficient plan refinement guided by \(P\), compared to unconstrained planning.

\begin{property}\label{prop:1}
Given an encoding of the answer set planning problem~$\Pi$ and a skeleton plan~$P$, a solution~$\tau$ of $(\Pi, P)$ always contains a plan for $\Pi$ and can be computed from the answer set of the encoding of $\Pi$ appending with the encoding for the skeleton plan~$P$.
\end{property}

\noindent  This holds because $P$ constrains the solution space of $\Pi$ without changing its underlying planning logic. Any answer set satisfying both still represents a valid plan for $\Pi$.

Property ~\ref{prop:1} enables skeleton plans to significantly reduce the search space, enhancing planning efficiency by decreasing the complexity from $O\bigl((S \times A)^N\bigr)$ to $O\bigl(m \times (S \times A)^{N/m}\bigr)$, where $S$ is the state space size, $A$ is the action space size, $N$ is the maximum number of timesteps, and $m$ is the number of skeleton steps. In our VirtualHome experiments, which involve numerous objects and relations, appropriate skeleton plans can accelerate answer set planning.

\subsection{Reduce Domains of Variables}
ASP solvers must ground program $\Pi$ into $\Pi|_{\HU{(\Pi)}}$ before computing answer sets. However, the size of $\Pi|_{\HU{(\Pi)}}$  can be exponentially larger than $\Pi$, creating a performance bottleneck. For example, VirtualHome task planning problems with hundreds of objects produce groundings exceeding 16MB at time 0, causing \emph{clingo} to timeout after 2 hours.

Our encoding $\Pi$ ensures safety through domain predicates (unary predicates given as facts). Each variable in a rule is restricted by a domain predicate in the rule's body. For instance, in rule $r$ at Line 14 of Listing \ref{lst:wash}, variable \texttt{O} is restricted by predicate \texttt{object}.

Since most objects are irrelevant for plan generation, we can omit them during answer set planning. This section discusses how to reduce the constant set for grounding while preserving solutions. We will define {\em admissible  reductions} and {\em safe reductions}, provide a sufficient condition for {\em admissible reduction}, and demonstrate its application in speeding up answer set planning.

\begin{definition}
Given an ASP program~$\Pi$ (with variables), a set~$\mathcal{D}$ of constants with $\mathcal{D} \subseteq \HU(\Pi)$ is an {\em admissible reduction} of $\Pi$, if
\begin{itemize}
\item for every answer set $I_\mathcal{D}$ of the ground program $\Pi|_{\mathcal{D}}$, there always exists an answer set $I$ of $\Pi$ such that $I_\mathcal{D}\subseteq I$.
\end{itemize}
\end{definition}
\begin{definition}
An {\em admissible reduction} $\mathcal{D}$ of $\Pi$ is also a {\em safe reduction} of $\Pi$, if
\begin{itemize}
\item for every answer set $I$ of $\Pi$, there always exists an answer set $I_\mathcal{D}$ of $\Pi|_{\mathcal{D}}$ such that $I_\mathcal{D}\subseteq I$.
\end{itemize}
\end{definition}
Note that for $\mathcal{D} \subseteq \HU(\Pi)$, while $\Pi|_{\mathcal{D}} \subseteq \Pi|_{\HU(\Pi)}$, $\mathcal{D}$ may neither be an admissible nor a safe reduction of ~$\Pi$. Consider program $\Pi'$:
\begin{align*}
&f(a). &  &f(b). & &g(a, b). &
&\gets g(x, y), f(x), f(y).
\end{align*}

where $\HU(\Pi')=\{a, b\}$, $x, y$ are variables appearing in $\Pi'$. With $\mathcal{D}=\{a\}$, $\Pi'|_{\mathcal{D}}$ has answer set $\{f(a), f(b), g(a, b)\}$, but $\Pi'$ has none. Thus $\{a\}$ is neither an admissible nor a safe reduction, while $\{a,b\}$ is both.

\begin{theorem}
Given a subset~$\mathcal{D}$ of $\HU(\Pi)$ for an ASP program~$\Pi$, deciding whether $\mathcal{D}$ is an admissible reduction or a safe reduction of $\Pi$ is coNP-hard.
\end{theorem}

\begin{proof}
We can construct an ASP problem~$\Pi$ from a 3-SAT problem by adding $\{ a \gets \Not a'. \ a'\gets \Not a.\}$ for each atom~$a$, without loss of generality, for each clause $\neg a \lor b \lor \neg c$ adding $\{ \gets a,\, \Not b,\, c.\}$, and adding $\{f(o_1).\ f(o_2).\ \gets f(x), f(y), x\neq y. \}$.
$\HU(\Pi)=\{o_1, o_2\}$, $\Pi|_{\HU(\Pi)}$ has no answer sets, $\Pi|_{\{o_1\}}$ has an answer set if{f} the 3-SAT problem is satisfiable. Then $\{o_1\}$ is an admissible or a safe reduction if{f} the 3-SAT problem is unsatisfiable.
\end{proof}

It is hard to identify admissible and safe reductions. In the application of answer set planning, rather than finding all solutions, we prefer to find an executable plan for the task. Then we consider the applications of {\em admissible reduction} in this paper.

With the help of the notions of loops and loop formulas, we provide a sufficient condition for {\em admissible reduction}.
\begin{definition}
\label{def:loop-admissible-reduction}
Given an ASP program~$\Pi$ (with variables), a set~$\mathcal{D}$ of constants with $\mathcal{D} \subseteq \HU(\Pi)$ is a {\em loop-admissible reduction} of $\Pi$, if
\begin{enumerate}
\item for every answer set $I_{\mathcal{D}}$ of $\Pi|_{\mathcal{D}}$, there exists an interpretation $I'$ such that $I_{\mathcal{D}} \cup I'$ satisfies rules in $\Pi|_{\HU(\Pi)}$ and loop formulas for every loop $L'$ of $\Pi$ with $L'\subseteq I'$, and
\item there does not exist a loop $L$ of $\Pi$ such that $L$ is not a loop of $\Pi|_{\mathcal{D}}$ and $L$ contains a loop $L'$ of $\Pi|_{\mathcal{D}}$ with $R^-(L', \Pi|_{\mathcal{D}}) \neq \emptyset$.
\end{enumerate}
\end{definition}

Intuitively, we require that the rules in \(\Pi|_{HU(\Pi)} \setminus \Pi|_{\mathcal{D}}\) and any newly introduced loop formulas can be satisfied by \(I_{\mathcal{D}} \cup I'\).

\begin{theorem}
Given a subset~$\mathcal{D}$ of $\HU(\Pi)$ for an ASP program~$\Pi$, if $\mathcal{D}$ is a loop-admissible reduction of $\Pi$, then $\mathcal{D}$ is an admissible reduction of $\Pi$.
\end{theorem}
\begin{proof}
Let $I_\mathcal{D}$ be an answer set of $\Pi|_{\mathcal{D}}$, from the definition of loop-admissible reduction, there exists the set $I'$ of groud atoms such that $I_\mathcal{D}\cup I'$ is a model of $\Pi$ and $I_\mathcal{D}\cup I'$ satisfies loop formulas of loops $L$ of $\Pi$ with $L\subseteq I_\mathcal{D}$ or $L\subseteq I'$.

To prove that $\mathcal{D}$ is an admissible reduction of $\Pi$, we need to show that $I_\mathcal{D} \cup I'$ is an answer set of $\Pi$. By Theorem~$\ref{them:1}$, it suffices to prove that $I_\mathcal{D} \cup I'$ satisfies $\LF(L, \Pi)$ for all loops $L$ of $\Pi$. Consider an arbitrary loop $L$ of $\Pi$. We analyze all possible cases:
\begin{enumerate}
\item Case $L \not\subseteq I_\mathcal{D} \cup I'$: By the the loop formula definition, the formula is vacuously true when the premise $\bigwedge_{A\in L} A$ is false.
\item Case $L \subseteq I_\mathcal{D}$: Since $I_{\mathcal{D}}$ is an answer set of $\Pi|_{\mathcal{D}}$ and for any rule $r\in R^-(L,\Pi|_{\mathcal{D}}) \subseteq R^-(L,\Pi)$, $I_{\mathcal{D}}$ satisfies $LF(L,\Pi)$.
\item Case $L \subseteq I'$: By condition $1$ of the loop-admissible reduction definition, $I_\mathcal{D} \cup I'$ satisfies $\LF(L, \Pi)$.
\item Case $L \cap I_\mathcal{D} \neq \emptyset \wedge L \cap I' \neq \emptyset \wedge L \subseteq I_\mathcal{D} \cup I'$:
We prove by contradiction that this case cannot occur. Suppose such a loop $L$ exists. Since $I_\mathcal{D}$ is an answer set of $\Pi|_\mathcal{D}$, by Theorem~$\ref{them:1}$, there must exist a loop $L' \subseteq L \cap I_\mathcal{D} \subseteq L$ that is at least a singleton, such that $R^-(L', \Pi|_\mathcal{D}) \neq \emptyset$. However, this contradicts condition $2$ of the loop-admissible reduction definition.
\end{enumerate}
So, for every loop $L$ of $\Pi$, $I_{\mathcal{D}}\cup I^{\prime}$ satisfies $LF(L,\Pi)$. Given that $I_{\mathcal{D}}\cup I^{\prime}$ is also a model of $\Pi$, by Theorem~\ref{them:1}, $I_{\mathcal{D}}\cup I^{\prime}$ is an answer set of $\Pi$.
\end{proof}
Notice that, a loop-admissible reduction $\mathcal{D}$ does not imply a safe reduction, as for a loop $L \subseteq I_\mathcal{D}$ in both $\Pi|_{\mathcal{D}}$ and $\Pi|_{\HU(\mathcal{D})}$, $\LF(L, \Pi|_{\mathcal{D}}) \supset \LF(L, \Pi|_{\HU(\mathcal{D})})$ but not vice versa.

\textbf{Application to Answer Set Planning.}
For a planning problem with skeleton plan $(\Pi, P)$, we define a loop-admissible reduction $\mathcal{D}^*$ as the minimal subset of the $HU(\Pi)$, that satisfies the following conditions: (1) $\mathcal{D} \subseteq \mathcal{D}^*$, where $\mathcal{D}$ is the initial set of relevant elements; and (2) $\mathcal{D}^*$ is closed under the binary relation $\mathcal{R} \subseteq HU(\Pi) \times HU(\Pi)$, such that $(o_1, o_2) \in \mathcal{R}$ and $o_1 \in \mathcal{D}^*$ imply $o_2 \in \mathcal{D}^*$. Formally, $\mathcal{D}^*$ can be expressed as:
\[
\mathcal{D}^* = \bigcup \left\{
\mathcal{S} \subseteq HU(\Pi) \ \middle|\
\begin{aligned}
& \mathcal{D} \subseteq \mathcal{S} \text{ and } \forall (o_1, o_2) \in \mathcal{R},\\
& \ o_1 \in \mathcal{S} \implies o_2 \in \mathcal{S}
\end{aligned}
\right\}.
\]
This expansion primarily affects binary relations (e.g., \texttt{on($o_1$,$o_2$)}, \texttt{in($o_1$,$o_2$)}, \texttt{close($o_1$,$o_2$)}). The resulting $\mathcal{D}^*$ satisfies both conditions for loop-admissible reduction because:
\begin{itemize}
\item Different domain predicates restrict variables in each rule of our encoding, ensuring $\Pi|_{\HU(\Pi)} \setminus \Pi|_{\mathcal{D}^*}$ and its loop formulas are satisfiable.
\item The construction of $\mathcal{D}^*$ ensures the second condition of Definition~\ref{def:loop-admissible-reduction} is met for all relevant loops.
\end{itemize}

Let $\Pi^*$ be the combination of encodings for $\Pi$ and $P$. We can compute answer sets of $\Pi^*|_{\mathcal{D}^*}$ to obtain the solution of $(\Pi, P)$.

\begin{property}\label{prop:2} Given the encoding $\Pi^*$ of a planning problem with a skeleton plan $(\Pi, P)$, the expanded set~$\mathcal{D}^*$ of constants is a loop-admissible reduction of $\Pi^*$.
A solution of $(\Pi, P)$ can be obtained from the answer set of $\Pi^*|_{\mathcal{D}^*}$.
\end{property}

\noindent This is guaranteed by the construction of $\mathcal{D}^*$, which ensures it contains all constants relevant to the planning problem while satisfying loop-admissible reduction conditions.

In our experiments, task planning in VirtualHome involves hundreds of objects. By reducing variable domains, we focus on dozens of objects and relations, greatly reducing ground program size. With appropriate skeleton plans, \emph{clingo} computation time was cut from over 2 hours to under 5 seconds.

\begin{table}[t]
\centering
\setlength{\tabcolsep}{0.1pt}
\small
\begin{tabular}{lccc}
    \toprule
    \multirow{1}{*}{Method} & LLM & $\overline{\textit{GCR}}$ & $R_{\text{exec}}$ \\
    \toprule
    \multirow{2}{*}{~\cite{huang2022language}}
    & GPT-3.5 & \phantom{0}6.27\,$\pm$\,\phantom{0}3.79\%  & \phantom{0}9.29\,$\pm$\,\phantom{0}8.43\%   \\
    & GPT-4o & \phantom{0}9.94\,$\pm$\,\phantom{0}6.10\%  & 11.90\,$\pm$\,10.48\% \\
    \midrule
    \multirow{2}{*}{~\cite{singh2023progprompt}}
    & GPT-3.5 & 20.42\,$\pm$\,\phantom{0}6.02\% & 16.67\,$\pm$\,\phantom{0}9.60\% \\
    & GPT-4o & 21.70\,$\pm$\,\phantom{0}8.04\% & 13.75\,$\pm$\,11.86\% \\
    \midrule
    \multirow{1}{*}{~\cite{mu2024embodiedgpt}$^*$} & Llama-3.1 & 26.73\,$\pm$\,\phantom{0}8.07\%& 32.71\,$\pm$\,22.01\% \\
    \midrule
    \multirow{3}{*}{\makecell{CLMASP-\\\;Clingo (ours)}}
    & Llama-3.1 & 13.69\,$\pm$\,\phantom{0}4.09\% & 81.04\,$\pm$\,15.36\% \\
    & GPT-3.5 & 37.07\,$\pm$\,\phantom{0}9.72\% & \textbf{89.59\,$\pm$\,\phantom{0}9.33\% }\\
    & GPT-4o & \textbf{41.90\,$\pm$\,10.75\%} & 86.99\,$\pm$\,11.32\% \\
    \bottomrule
\end{tabular}
\caption {\textbf{Comparison of CLMASP-Clingo with other baselines.} Results using (Clingo) Full are shown in Table~\ref{tab:ablation_study_CLMASP}. Method marked with $^*$ use a fine-tuned LLM.
}
\label{tab:baseline_comparsion}
\end{table}

\begin{table*}[!tb]
\centering
\setlength{\tabcolsep}{4.35pt}
\begin{tabular}{@{}ll|ccccc|ccccc|cc@{}}
\toprule
\multirow{3}{*}{Solver} & \multirow{3}{*}{Model}
    & \multicolumn{5}{c|}{$T_\text{ASP}$}
    & \multicolumn{5}{c|}{$R_\text{sol}$}
    & \multicolumn{2}{c}{$L_{\tau_f}$} \\
\cmidrule(r){3-7} \cmidrule(l){8-12} \cmidrule(l){13-14}
 & & -S-R & -R & -S & Full & $\Delta$
    & -S-R & -R & -S & Full & $\Delta$
    & -S & Full \\
\midrule
\multirow{3}{*}{DLV2}
  & Llama-3.1 & \textgreater2h & \textgreater2h & 39.37s & 23.33s & -40.7\%
              & 0 & 0 & 7.4\% & 65.4\% & +58.0\%
              & (4.75/6/13) & (6.08/\phantom{0}7/29) \\
  & GPT-3.5  & \textgreater2h & \textgreater2h & 37.24s & 24.38s & -34.5\%
              & 0 & 0 & 7.4\% & 93.3\% & +85.9\%
              & (4.75/6/13) & (7.53/10/30) \\
  & GPT-4o   & \textgreater2h & \textgreater2h & 38.75s & 29.29s & -24.4\%
              & 0 & 0 & 7.4\% & 91.4\% & +84.0\%
              & (4.75/6/13) & (8.23/11/30) \\
\midrule
\multirow{3}{*}{Clingo}
  & Llama-3.1 & \textgreater2h & \textgreater2h & 32.06s & \phantom{0}4.16s & -87.1\%
              & 0 & 0 & 7.4\% & 84.0\% & +76.6\%
              & (4.75/6/13) & (5.37/\phantom{0}7/24) \\
  & GPT-3.5  & \textgreater2h & \textgreater2h & 30.52s & \phantom{0}4.63s & -84.8\%
              & 0 & 0 & 7.4\% &  93.7\% & +86.3\%
              & (4.75/6/13) & (7.17/\phantom{0}9/28) \\
  & GPT-4o   & \textgreater2h & \textgreater2h & 31.17s & 11.20s & -64.1\%
              & 0 & 0 & 7.4\% & 91.4\% & +84.0\%
              & (4.75/6/13) & (8.20/11/28) \\
\bottomrule
\end{tabular}
\caption{\textbf{Ablation study on the skeleton plan and loop-admissible reduction techniques in CLMASP}. Each variant removes the corresponding technique from CLMASP (Full): removing loop-admissible reduction is denoted as (-R), removing skeleton plan is denoted as (-S), and removing both of them is denoted as (-S-R). The table also shows the relative improvement of Full over -S, denoted as $\Delta$.}
\label{table:asptime_exp}
\end{table*}

\begin{table*}[!t]
\centering
\setlength{\tabcolsep}{3pt}
\begin{tabular}{cccccccc}
\toprule
\multicolumn{2}{c}{} & \multicolumn{6}{c}{\textbf{Method}} \\
\cmidrule(lr){3-8}
\multicolumn{2}{c}{LLM} & -SGR-ASP & -ASP & (DLV2)-SGR & (Clingo)-SGR  & (DLV2) Full & (Clingo) Full \\
\midrule
\multirow{2}{*}{Llama-3.1}
& $\overline{\textit{GCR}}$ & \phantom{0}3.83\,$\pm$\,\phantom{0}1.48\% & 11.32\,$\pm$\,\phantom{0}3.01\% & \phantom{0}8.28\,$\pm$\,\phantom{0}3.53\% & \phantom{0}8.61\,$\pm$\,\phantom{0}3.51\% & 11.41\,$\pm$\,\phantom{0}3.81\% & \textbf{13.69\,$\pm$\,\phantom{0}4.09\%} \\

& $R_\text{exec}$ & \phantom{0}3.72\,$\pm$\,\phantom{0}3.58\% & 27.14\,$\pm$\,19.77\% & 39.41\,$\pm$\,23.88\% & 42.01\,$\pm$\,24.36\% & 62.45\,$\pm$\,23.45\% & \textbf{81.04\,$\pm$\,15.36\%} \\

\midrule
\multirow{2}{*}{GPT-3.5}
& $\overline{\textit{GCR}}$ & 14.05\,$\pm$\,\phantom{0}6.35\% & 23.69\,$\pm$\,\phantom{0}8.07\% & 35.93\,$\pm$\,\phantom{0}9.98\% & 35.35\,$\pm$\,\phantom{0}9.92\% & \textbf{37.18\,$\pm$\,\phantom{0}9.89\%} & 37.07\,$\pm$\,\phantom{0}9.72\%\\
& $R_\text{exec}$ & \phantom{0}8.55\,$\pm$\,\phantom{0}7.82\% & 11.15\,$\pm$\,\phantom{0}9.91\% & 85.87\,$\pm$\,12.13\% & 86.99\,$\pm$\,11.32\% & 88.48\,$\pm$\,10.20\% & \textbf{89.59\,$\pm$\,\phantom{0}9.33\%} \\
\midrule
\multirow{2}{*}{GPT-4o}
& $\overline{\textit{GCR}}$ & 22.55\,$\pm$\,\phantom{0}7.41\% & 28.62\,$\pm$\,\phantom{0}8.06\% & 42.40\,$\pm$\,10.68\% & 41.76\,$\pm$\,10.94\% & \textbf{42.34\,$\pm$\,10.84\%} & 41.90\,$\pm$\,10.75\% \\
& $R_\text{exec}$ & 11.52\,$\pm$\,10.20\% & 15.24\,$\pm$\,12.92\% & 87.36\,$\pm$\,11.04\% & 86.25\,$\pm$\,11.86\% & \textbf{87.73\,$\pm$\,10.76\%} & 86.99\,$\pm$\,11.32\% \\
\bottomrule
\end{tabular}
\caption{\textbf{Ablation study on the Semantic Grounding and Refinement (SGR) and ASP solving steps in CLMASP}. Each variant removes the corresponding step from CLMASP (Full): removing Semantic Grounding and Refinement is denoted as (-SGR), removing ASP solving is denoted as (-ASP), and removing both is denoted as (-SGR-ASP). The ASP solver used is indicated in parentheses.}
\label{tab:ablation_study_CLMASP}
\end{table*}

\section{Couple LLMs with Answer Set Planning}
To effectively integrate LLM common sense with ASP's reasoning capabilities for real-world planning tasks, we propose CLMASP as a two-phase planning framework, as illustrated in Figure~\ref{fig:workflow}. This framework first leverages the LLM for rough planning and grounding, followed by ASP to elaborate the skeleton plan into a final solution.

\subsection{Generating a Skeleton Plan by LLMs}

\textbf{Initial Plan Generation.}
Given a task description, primitive actions, environmental object categories, and three planning examples, LLMs generate an initial plan, $\tau_s^0$. This plan uses object categories instead of specific IDs due to LLM input capacity limitations.

\textbf{Semantic Grounding and Refinement (SGR).}
The SGR step refines the initial plan $\tau_s^0$ by correcting invalid or ambiguous terms (e.g., ``walk kitchencounter'' to ``walk kitchen'' because ``kitchencounter'' is absent).
It computes cosine similarities ($\cos(\theta) = \frac{\vec{a} \cdot \vec{b}}{|\vec{a}| |\vec{b}|}$) between plan element embeddings ($a$) and environmental component embeddings ($b$).
Invalid terms are then substituted with the closest valid environmental actions or objects. This yields a semantically refined plan, $\tau_s$, ready for ASP reasoning and execution.

\subsection{Enhancing a Skeleton Plan by ASP}
The skeleton plan $\tau_s$ is refined using ASP to establish action dependencies and specify object locations. First, reusable ASP codes encode robotic action models (e.g., cause-effect relationships). ASP's initial state is then derived from a global environment map---a directed acyclic graph of entity nodes and relationship edges (e.g., \texttt{is(1,character)}, \texttt{state(7,dirty)}, \texttt{relation(in,5,2)}). Finally, a Python module translates $\tau_s$ into ASP rules, treating each subtask as a temporally ordered goal to produce the final plan $\tau_f$. This process is detailed in Section~\ref{sec:ImproveEfficiency}.

\section{Evaluation}
\subsection{Experiment Setup}
\textbf{Simulator and Dataset.}

Experiments utilize the VirtualHome (VH) simulator v2.3.0, where an agent interacts with 250–300 household objects using 42 actions. These objects, linked by approximately 3,000 relationships, form an environment abstracted as a grounded directed acyclic graph. As shown in Fig.~\ref{fig:workflow}, task inputs comprise a natural language description (\textit{task}), the environment as a serialized dictionary (\textit{Env.}), and structured possible actions (\textit{Act.}). Each task instance includes a reference plan—the ground truth sequence of verb-object pairs that accomplishes the \textit{task}. (e.g., [WALK] \textless wardrobe\textgreater, [GRAB] \textless clothes\textgreater, [PUTIN] \textless clothes\textgreater  \textless clothesbin\textgreater for ``put clothes in bin"). For evaluation, we sample 269 of 292 VH task instances and reserve 23 for prompting examples.

\textbf{LLMs and Solvers.}
Evaluations of CLMASP and baselines utilize the LLMs \emph{gpt-3.5-turbo-1106}~\cite{brown2020language}, \emph{gpt-4o-2024-08-06}~\cite{hurst2024gpt}, and \emph{Llama-3.1-8B}~\cite{meta_llama_3}. For its ASP solver, CLMASP employed \emph{clingo} 5.7.1~\cite{gebser2019multi} and \emph{DLV2}~\cite{calimeri2022asp}. The SGR step in CLMASP used the \emph{text-embedding-ada-002} embedding model.
To ensure fairness, all experimental results in this paper are obtained from actual measurements rather than citations.

\textbf{Metrics.}
Metrics include \emph{ASP Planning Time} ($T_\text{ASP}$), the \emph{ASP runtime} (Fig.~\ref{fig:workflow}, lower part); \emph{Solvability Rate} ($R_\text{sol}$), the proportion of instances where the final plan $\tau_f$ is successfully solved; \emph{Executability Rate} ($R_\text{exec}$), the portion of $\tau_f$ successfully executed in VH; and \emph{Final Plan Length Stats} ($L_{\tau_f}$), covering average, median, and maximum $\tau_f$ length.
We also measure \emph{Mean Goal Condition Recall} ($\overline{\textit{GCR}}$), averaged across all instances. For each instance, \textit{GCR} is the proportion of goal states and relations achieved:
\begin{equation*} \textit{GCR} = 1 - \frac{|(\mathcal{D}_{\text{goal}} - \mathcal{D}_{\text{initial}}) - (\mathcal{D}_{\text{method}} - \mathcal{D}_{\text{initial}})|}{|\mathcal{D}_{\text{goal}} - \mathcal{D}_{\text{initial}}|}, \end{equation*}
where $\mathcal{D}_{\text{initial}}$, $\mathcal{D}_{\text{method}}$, and $\mathcal{D}_{\text{goal}}$ are the sets of object states and relations at the initial state, after method application, and in the goal state, respectively.

\subsection{Results and Analysis}
To verify CLMASP's effectiveness, we compare it with three baselines. As shown in Table~\ref{tab:baseline_comparsion}, CLMASP's $R_{\text{exec}}$ (90\%) and $\overline{\textit{GCR}}$ (42\%) significantly surpass the best baseline (33\% and 27\%, respectively), indicating CLMASP effectively improves executability and goal condition fulfillment of LLM-based planning in large domains.

To assess the contribution of CLMASP's two ASP planning acceleration techniques, we conduct an ablation study. As shown in Table~\ref{table:asptime_exp}, removing loop-admissible reduction (-R) results in $T_\text{ASP}$ exceeding 2 hours; including it drops solving time below 40 seconds. Adding the skeleton plan technique further improves solving efficiency by 24\%-87\% and significantly increases $R_\text{sol}$ by approximately 58\%-86\%. The method is effective for $L_{\tau_f}$ across different LLMs and solvers. Therefore, these two techniques, especially the reduction technique, are crucial for accelerating ASP planning and key to its practicality.

To verify the necessity of CLMASP's dual-phase approach, we also conduct an ablation study. As shown in Table~\ref{tab:ablation_study_CLMASP}, removing the ASP solving step leads to plans with only 3.7\%-27.1\% $R_\text{exec}$. In contrast, removing SGR still maintains 39\%-87\% $R_\text{exec}$. However, only when both components are included does the method achieve 89.59\% $R_\text{exec}$ and around 40\% average $\overline{\textit{GCR}}$. Thus, the ASP solving step is CLMASP's core phase, while SGR serves as a powerful enhancement to fully exploit its potential by correcting the grounding and utilization of objects.

A horizontal comparison in Table~\ref{table:asptime_exp} and \ref{tab:ablation_study_CLMASP} reveals that while clingo and DLV2 offer similar ASP solving capabilities, clingo is more efficient. Regarding LLMs, their performance mainly dictates skeleton plan quality, which impacts $\overline{\textit{GCR}}$ more significantly (sometimes negatively) than $R_\text{exec}$.

\section{Conclusion}
This paper defines the planning problem with a skeleton plan and encodes the ASP program to refine the skeleton plan to accomplish the task. We introduce {\em admissible and safe reductions} that preserve solutions, and to address the challenge of identifying them, we propose a sufficient condition for {\em admissible reductions} using loops and loop formulas. Using these speedup techniques, we present CLMASP, an approach coupling LLMs with ASP for robotic task planning. Experiments on the VirtualHome platform demonstrate CLMASP's improvements in both computational efficiency and plan executable rates.

\appendix

\section*{Acknowledgments}
This work was supported by the National Natural Science Foundation of China (No. 62332016), the Key Research Program of Frontier Sciences, CAS (No. ZDBS-LY-JSC001), the National Key R\&D Program of China (No. 2023YFB4704500), and the Hunan Province Major Scientific and Technological Project (No. 2024QK200).

\section*{Contribution Statement}
The first two authors contributed equally to this work.

\bibliographystyle{named}
\bibliography{ijcai25}

\end{document}